\pdfoutput=1
\PassOptionsToPackage{table}{xcolor} 

\documentclass[11pt]{article}

\usepackage{EMNLP2023}

\usepackage{times}
\usepackage{latexsym}
\usepackage{amsmath}
\usepackage{bm}
\usepackage{times}
\usepackage{latexsym}
\usepackage{amsmath}
\usepackage{graphicx}
\usepackage{bm}
\usepackage{amsfonts,amssymb}
\usepackage{bbm}
\usepackage{booktabs}
\usepackage{booktabs,siunitx}
\usepackage{multirow}
\usepackage{subfigure}
\usepackage{parskip}
\usepackage{graphicx}
\usepackage{subfigure}
\usepackage{graphicx}
\usepackage[T1]{fontenc}
\usepackage[utf8]{inputenc}
\usepackage{microtype}
\usepackage{array}
\usepackage{booktabs}
\usepackage{multirow}
\usepackage{algorithm}
\usepackage{algorithmic}

\usepackage[T1]{fontenc}

\usepackage[utf8]{inputenc}

\usepackage{microtype}

\usepackage{inconsolata}

\title{Knowledge-Augmented Multimodal Clinical Rationale Generation for Disease Diagnosis with Small Language Models}

\author{Shuai Niu$^1$, Jing Ma$^1$, Hongzhan Lin$^1$, 
    Liang Bai$^2$, Zhihua Wang$^3$, \\ \textbf{Yida Xu$^1$,} \textbf{Yunya Song$^4$,} and \textbf{Xian Yang$^5$}\thanks{* This is the corresponding author.}
    \\
    Hong Kong Baptist University$^1$,
    Shanxi University$^2$ ,\\
    Shanghai Institute for Advanced Study of Zhejiang University$^3$,\\
    Hong Kong University of Science and Technology$^4$, The University of Manchester$^5$\\
    \texttt{\{cssniu,majing\}@comp.hkbu.edu.hk,  xian.yang@manchester.ac.uk} \\
}

\begin{document}
\maketitle
\begin{abstract}
Interpretation is critical for disease diagnosis, but existing models struggle to balance predictive accuracy with human-understandable rationales. While large language models (LLMs) offer strong reasoning abilities, their clinical use is limited by high computational costs and restricted multimodal reasoning ability. Small language models (SLMs) are efficient but lack advanced reasoning for integrating multimodal medical data. In addition, both LLMs and SLMs lack domain knowledge for trustworthy reasoning. Therefore, we propose ClinRaGen, enhancing SLMs by leveraging LLM-derived reasoning ability via rationale distillation and domain knowledge injection for trustworthy multimodal rationale generation. Key innovations include a sequential rationale distillation framework that equips SLMs with LLM-comparable multimodal reasoning abilities, and a knowledge-augmented attention mechanism that jointly unifies multimodal representation from time series and textual data in the same encoding space, enabling it to be naturally interpreted by SLMs while incorporating domain knowledge for reliable rationale generation. Experiments on real-world medical datasets show that ClinRaGen achieves state-of-the-art performance in disease diagnosis and rationale generation, demonstrating the effectiveness of combining LLM-driven reasoning with knowledge augmentation for improved interpretability.
\end{abstract}

\section{Introduction}

\begin{figure}[t]
\centering
\centerline{\includegraphics[scale=0.31 ]{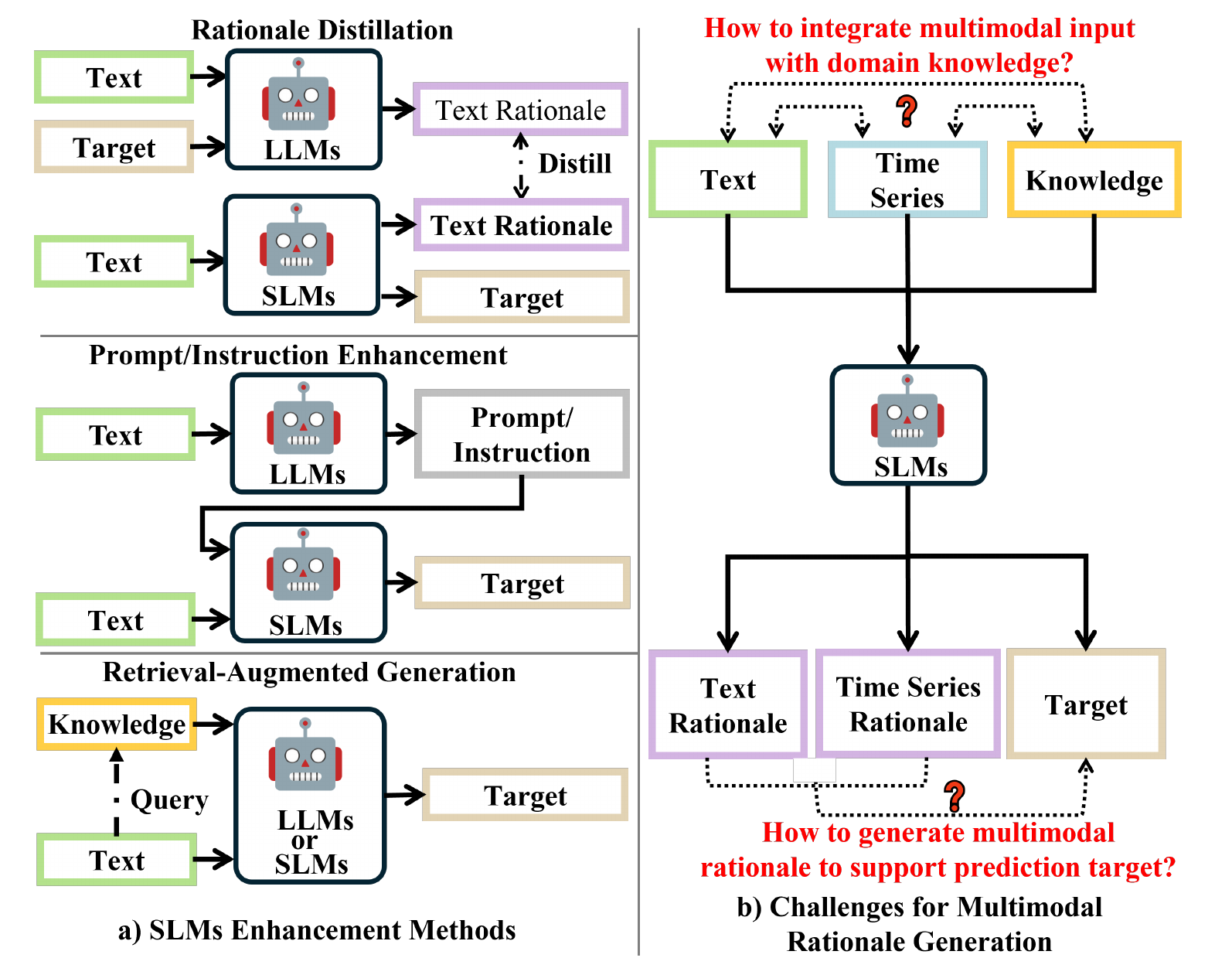}}
\caption{ Existing SLM enhancement methods and challenges in multimodal rationale generation.}
\label{Intro}
\end{figure}

The widespread adoption of electronic health records (EHRs) has transformed deep learning applications in healthcare by providing diverse data modalities, including medical notes, laboratory (lab) test results, and clinical events. These multimodal inputs are crucial for disease diagnosis, mortality prediction, and drug discovery \cite{niu2024enhancing,niu2024ehr,laghuvarapu2024codrug}. Large language models (LLMs) have recently demonstrated strong diagnostic performance and reasoning capabilities through techniques such as prompt learning and Chain-of-Thought (CoT) reasoning \cite{wei2022chain,singhal2023large,chen2023meditron}. However, despite these advancements, LLMs face significant challenges in real-world clinical deployment due to high computational costs, {the need for external domain-specific data integration}, and difficulties in processing multimodal inputs, particularly numerical time-series lab tests. More critically, LLMs lack the ability to generate clinically grounded multimodal rationales, limiting their interpretability in medical decision-making.

Small language models (SLMs) have emerged as a computationally efficient alternative, benefiting from recent advancements in rationale distillation, prompt learning, and retrieval-augmented generation (RAG) \cite{hsieh2023distilling,kang2024knowledge,kwon2024large}. As shown in Figure \ref{Intro}a, these methods enable SLMs to inherit LLM-driven reasoning abilities, improve generalization through instruction-based adaptation, or leverage RAG for more reliable outputs. However, as illustrated in Figure \ref{Intro}b, these approaches still suffer from two fundamental challenges. The first challenge is that they struggle to effectively integrate multimodal inputs with structured domain knowledge, as most methods focus on single-modality data (e.g., text-based rationales) rather than jointly processing textual and time series EHR data \cite{shi2024medadapter,sohn2024rationale}. The second challenge is that they fail to provide coherent multimodal rationales that align with clinical decision-making, as rationale generation often remains text-centric and lacks interpretability across different data modalities.

{To bring the best of both worlds, we propose ClinRaGen, a knowledge-augmented framework for multimodal clinical rationale generation. ClinRaGen enhances SLMs' trustworthy multimodal reasoning capabilities from two aspects. First, it transfers LLM-derived reasoning to SLMs through a sequential rationale distillation paradigm. Second, unlike approaches that rely solely on LLM-generated rationales \cite{kwon2024large} or resource-intensive RAG \cite{kang2024knowledge}, we propose a knowledge-augmented attention mechanism that achieves dual functionality: {Efficient integration of external medical knowledge} to enable multimodal rationale generation grounded in clinical validity, ensuring the production of clinically meaningful explanations;  {Unification of time-series and textual EHRs within a shared encoding space}, thereby enhancing multimodal representation learning and facilitates interpretable decision-making.}

The main contributions of this paper are:
\begin{itemize}

    \item We propose ClinRaGen, a multimodal framework that transfers LLM reasoning capabilities into SLMs for disease diagnosis and clinical rationale generation, achieving both accuracy and interpretability.
    \item We introduce a knowledge-augmented attention mechanism that jointly encodes time-series EHRs into clinical textual representations while injecting domain knowledge, significantly improving multimodal rationale reliability and accuracy.
    \item State-of-the-art performance in disease diagnosis and rationale generation, validated through extensive experiments on benchmark EHR datasets~\cite{johnson2016mimic,johnson2023mimic}.
\end{itemize}

\section{Related Work}

Recent advancements in highly effective attention mechanisms \cite{vaswani2017attention, niu2021label, dao2023flashattention,yang2024hyperspectral}, large-scale  CoT datasets \cite{cobbe2021training,zhong2023agieval}, improved computational resources have enabled significant progress in Natural Language Processing (NLP), with improved training methodologies fueling the development of LLMs \cite{touvron2023llama, achiam2023gpt}. In healthcare, LLMs have been applied to clinical question answering and diagnostic reasoning \cite{singhal2023large, yang2022gatortron}. While effective in text-based tasks, these models struggle to generate clinically grounded multimodal rationales. Medical-specific LLMs \cite{chen2023meditron, zhang2023alpacare} mitigate this issue through domain adaptation, but their high computational costs and reliance on large-scale training data limit scalability.

To improve efficiency, rationale distillation transfers LLM-derived reasoning ability to SLMs, reducing computational overhead while preserving interpretability \cite{lin2023beneath, hsieh2023distilling, ho2023large, kang2024knowledge}. Chain-of-thought prompting further enhances SLM reasoning capabilities \cite{wei2022chain}. However, most distillation approaches remain text-centric and lack robust multimodal EHRs integration \cite{kang2024knowledge, ho2023large}. RAG has been explored to improve rationale reliability by incorporating external knowledge, yet retrieval latency and adaptability remain key challenges \cite{jiang2024reasoning}. Despite these advancements, multimodal rationale generation remains an open challenge. Current models struggle to fuse textual, time-series, and medical knowledge into coherent clinical rationales.

\begin{figure*}[h]
\centering
\centerline{\includegraphics[scale=0.42]{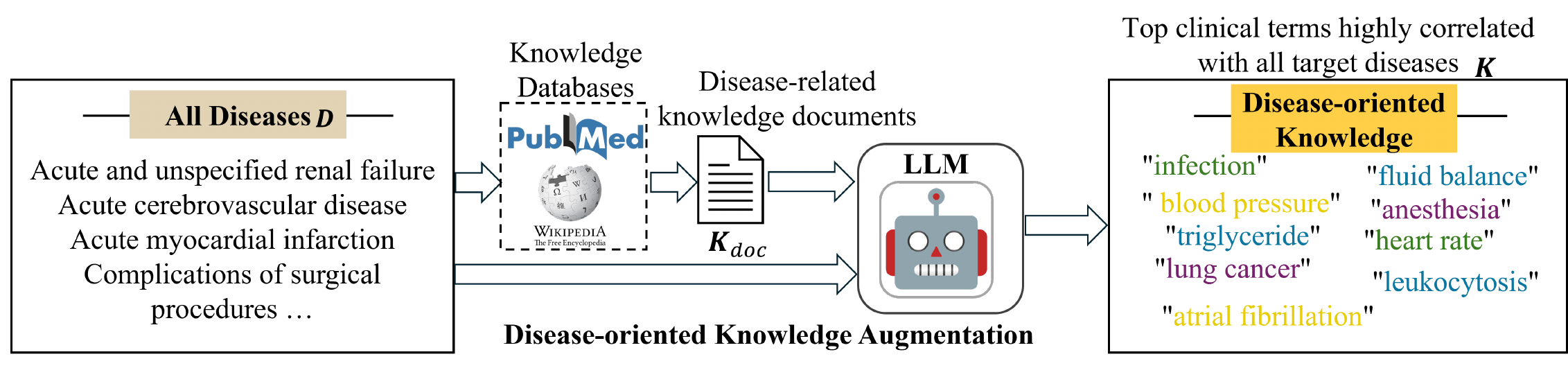}}
\caption{Knowledge augmentation in ClinRaGen. Given diagnosed diseases $\bm{D}$, relevant descriptions $\bm{K}_{doc}$ are retrieved and processed by an LLM to extract key clinical terms $\bm{K}$, enhancing multimodal rationale generation. }
\label{model2}
\end{figure*}

\section{Methodology}
We introduce ClinRaGen, a knowledge-augmented framework designed to enhance disease diagnosis and clinical rationale generation in SLMs by integrating LLM-derived reasoning and structured domain knowledge. ClinRaGen bridges the gap between large-scale medical knowledge and efficient multimodal reasoning, enabling the generation of two types of rationales: 1). medical note-based rationales ($\bm{R}^m$) and 2). lab test-based rationales ($\bm{R}^t$) from medical notes  ($\bm{M}$), time-series lab test results ($\bm{T}$), and disease-specific knowledge ($\bm{K}$).

ClinRaGen consists of two key components for the knowledge distillation: Knowledge Retrieval and LLM-Guided Rationale Generation (Section 3.1), which collects domain knowledge and generates LLM-derived rationales as distillation data for subsequent model training, and Multimodal Rationale Distillation (Section 3.2), which progressively integrates structured knowledge to enhance multimodal reasoning in SLMs.

\subsection{Knowledge Retrieval and LLM-Guided Rationale Generation}
This step focuses on gathering domain knowledge and leveraging LLMs to generate structured rationales. The generated rationales serve as distillation targets for training SLMs in later stages. This ensures that SLMs receive high-quality, structured reasoning data to develop robust multimodal reasoning capabilities.

\subsubsection{Collecting Domain-Specific Medical Knowledge}
LLMs encode extensive medical knowledge but are computationally expensive and impractical for direct deployment. Meanwhile, SLMs such as Flan-T5 and  Flan-PaLM~\cite{chung2024scaling} are computationally efficient but lack sufficient domain-specific expertise to perform complex medical reasoning~\cite{kang2024knowledge, ho2023large}. To bridge this gap, ClinRaGen retrieves relevant medical knowledge from external sources and structures it for integration into SLM training.

As shown in Figure~\ref{model2}, ClinRaGen collects disease-related documents $\bm{K}_{doc}$ from PubMed\footnote{https://pubmed.ncbi.nlm.nih.gov/} and Wikipedia\footnote{https://www.wikipedia.org/}, extracting key medical terms using LLM-based processing to construct a structured knowledge base $\bm{K}$:
\begin{equation} 
\bm{K} = \underset{\bm{K}'}{\operatorname{argmax}} P_{LLM}(\bm{K}' \mid \bm{D}, \bm{K}_{doc}). 
\end{equation}
This structured knowledge base is not used directly by the SLM during inference but instead supports rationale generation in the next step. The retrieval and extraction process iterates until a stable set of key medical terms is obtained.

\subsubsection{Generating Rationales for Distillation}
ClinRaGen employs LLMs to generate structured rationales that serve as distillation targets for SLM training. Unlike direct knowledge retrieval\cite{kang2024knowledge,jiang2024reasoning}, this step synthesizes structured explanations that explicitly link medical knowledge with clinical decision-making, enabling SLMs to internalize complex reasoning patterns during later training stages.
To construct high-quality rationale data, we collaborated with clinicians to curate representative EHR samples and formulate corresponding gold-standard rationales $\bm{O}$. These rationales guide the LLM in generating structured explanations, ensuring that the distilled knowledge supports multimodal reasoning.
To improve LLM comprehension of numerical lab test data, we applied anomaly detection~\cite{vinutha2018detection} and designed structured prompts that convert numerical values into interpretable textual explanations $\bm{T}^{*}$ (see Appendix A.1).

Figure~\ref{model1} illustrates the multimodal rationale generation process. ClinRaGen first generates rationales ($\bm{R}^m$) based on medical notes:
\begin{equation} 
\bm{R}^m = \underset{\bm{R}'}{\operatorname{argmax}} P_{LLM}(\bm{R}' \mid \bm{M}, \bm{D}, \bm{O}). 
\end{equation}
Then, lab test-based rationales ($\bm{R}^t$) are generated using insights from both medical notes, time series anomalies, and the generated note-based rationales:
\begin{equation} 
\bm{R}^t = \underset{\bm{R}'}{\operatorname{argmax}} P_{LLM}(\bm{R}' \mid \bm{M}, \bm{T}^{*}, \bm{D}, \bm{O}, \bm{R}^m). 
\end{equation}
These LLM-generated rationales form the foundation of the subsequent distillation process (detailed in Section 3.2) and enable SLMs to learn structured, multimodal reasoning efficiently.
For further details on data processing and prompt engineering, refer to Appendix A.2.

\begin{figure}[t]
\centering
\centerline{\includegraphics[scale=0.30]{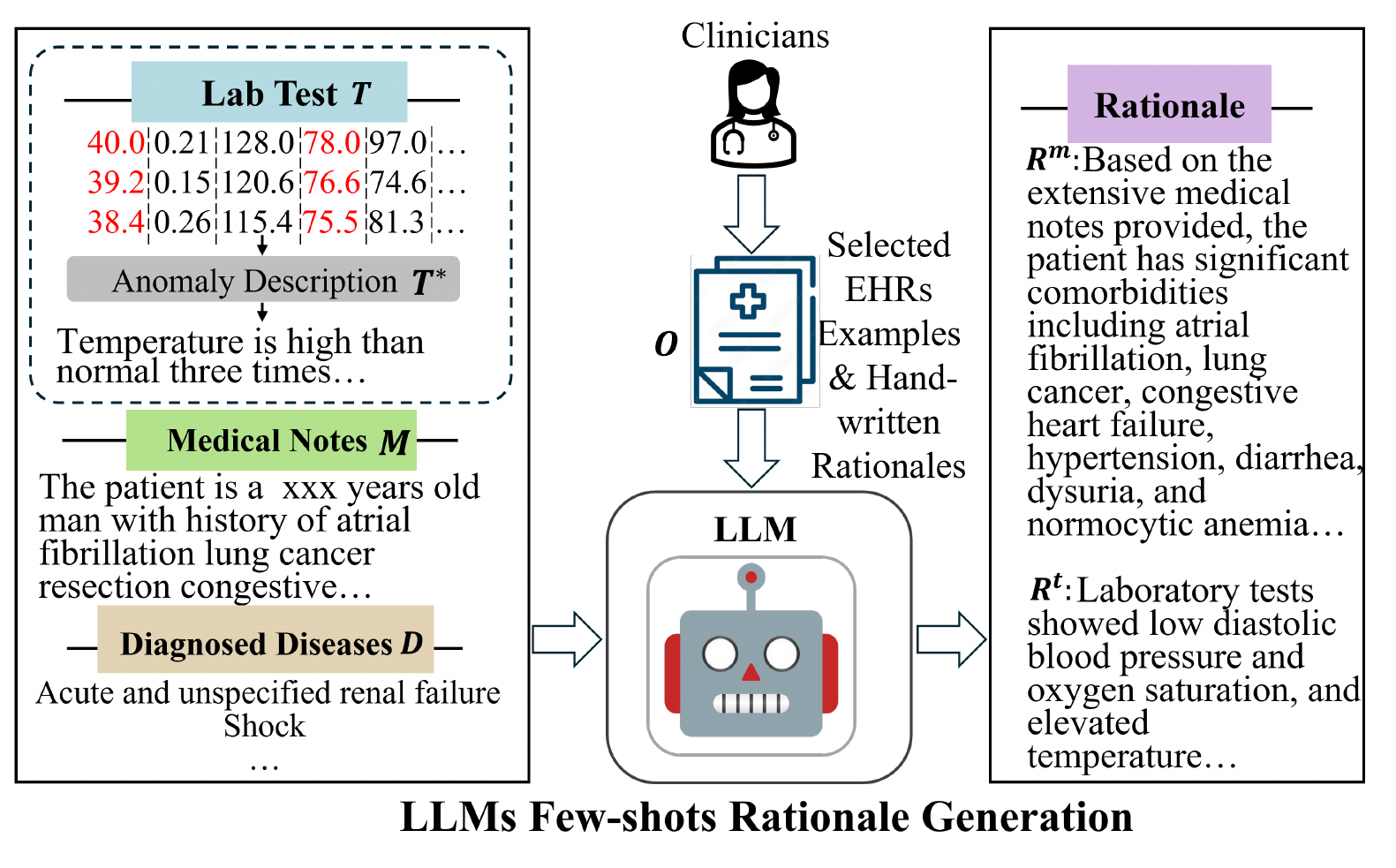}}
\caption{LLM-based clinical rationale generation. Medical notes ($\bm{M}$), lab test results ($\bm{T}$ and $\bm{T}^*$), diagnosis ($\bm{D}$), and clinicians provide examples ($\bm{O}$) are used to produce medical note-based ($\bm{R}^m$) and lab test-based ($\bm{R}^t$) rationales.}
\label{model1}
\end{figure}

\begin{figure*}[t]
\centering
\centerline{\includegraphics[scale=0.365]{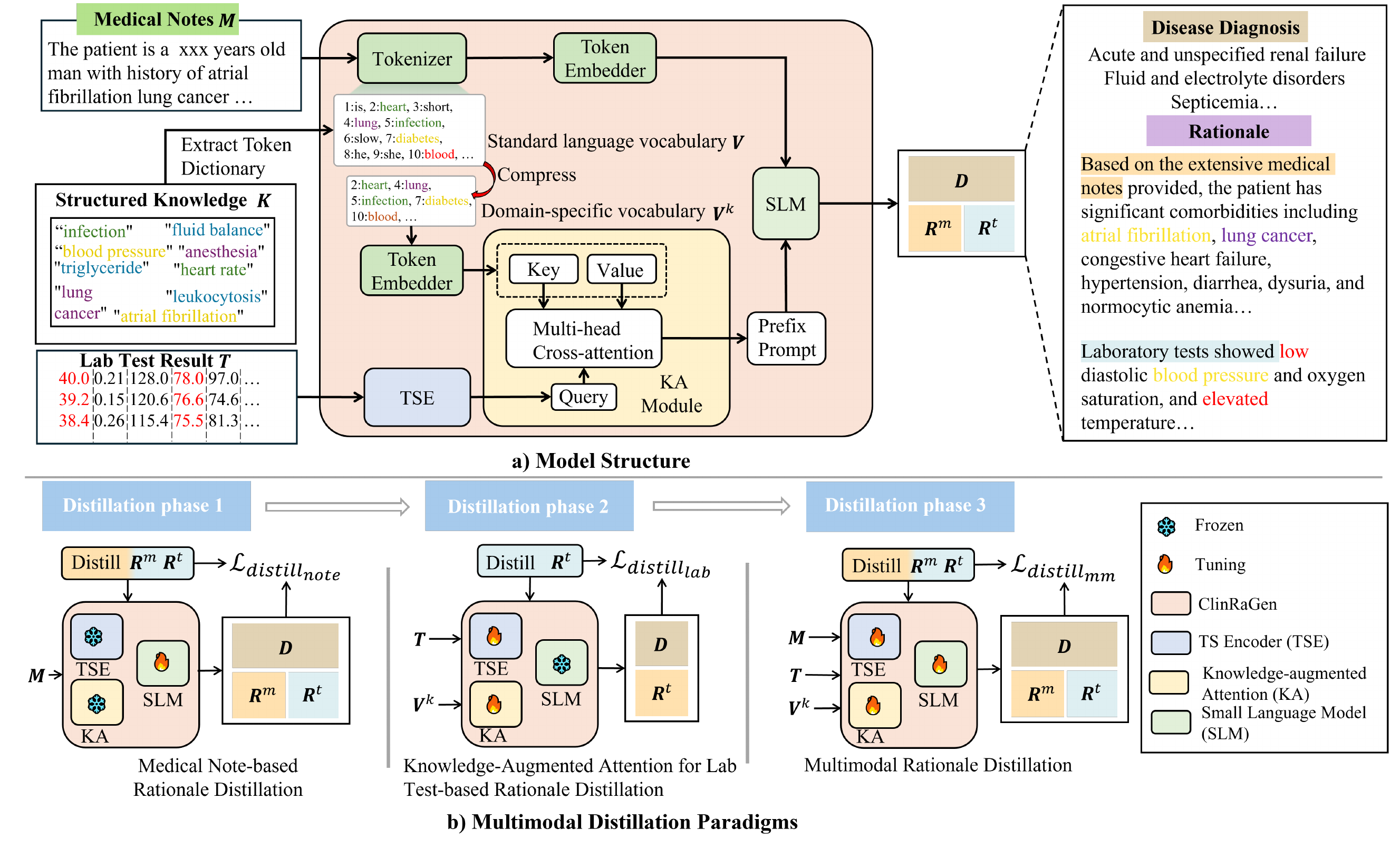}}
\caption{
Overview of ClinRaGen. (a) Model structure comprising a time series encoder, a knowledge-augmented attention module, and an SLM. (b) Three-phase rationale distillation: Medical Note-based Rationale Distillation, Knowledge-augmented attention for Lab Test-based Rationale Distillation, and Multimodal Rationale Distillation.
}
\label{model3}
\end{figure*}

\subsection{Multimodal Rationale Distillation} 
Figure~\ref{model3}a presents the ClinRaGen framework, which comprises a Time Series Encoder for processing numerical lab test data, a knowledge-augmented attention module for integrating structured domain knowledge, and an SLM for generating disease diagnoses and structured clinical multimodal rationales. The framework enables progressive multimodal reasoning by leveraging structured knowledge and sequential learning mechanisms.

As illustrated in Figure~\ref{model3}b, ClinRaGen employs a three-phase rationale distillation paradigm that systematically integrates textual, numerical, and structured domain knowledge. The first phase distils medical note-based rationales, allowing the SLM to develop a foundational understanding of textual clinical information. The second phase introduces knowledge-augmented attention, aligning numerical lab tests with structured medical knowledge to distil lab test-based rationales. The final phase fully integrates textual and numerical inputs, enabling the SLM to generate clinically coherent multimodal rationales to support disease diagnosis.

\subsubsection{Phase 1: Rationale Distillation from Medical Notes}
In the first phase, the SLM is trained exclusively on medical notes $\bm{M}$ to establish a foundational understanding of clinical reasoning. This stage enables the model to generate disease diagnoses $\bm{D}$ while also producing medical note-based rationales $\bm{R}^m$ and lab test-based rationales $\bm{R}^t$. By learning to extract meaningful insights from structured textual data, the SLM develops its initial ability to infer clinical relationships. The model is trained using a language model generation objective:
\begin{equation}
\mathcal{L}_{note}(\theta) = \mathbb{E}[-\log P_{SLM_{\theta}}(\bm{D}, \bm{R}^m, \bm{R}^t \mid \bm{M})],
\end{equation}
where $\theta$ represents the trainable parameters of the SLM. This phase not only enables the model to internalize explicit diagnostic reasoning from medical notes but also allows it to implicitly capture latent patterns associated with lab test results, laying the groundwork for multimodal integration in subsequent phases.

\subsubsection{Phase 2: Knowledge Injection and Time-Series Rationale Distillation} 

To enable the SLM to effectively interpret numerical lab test data and generate time-series-based rationales ($\bm{R}^t$) that support disease diagnosis ($\bm{D}$), we introduce a Knowledge-augmented Attention (KA) Module. This mechanism integrates domain-specific medical knowledge into the reasoning process, with the aim of enhancing the model's focus, shifting it away from high-frequency but clinically irrelevant tokens towards clinically significant information. Consequently, this improved attention allows the model to produce more clinically coherent and robust multimodal rationales.

We first use a Time Series Encoder (TSE) to encode raw lab test values $\bm{T}$ into structured hidden embeddings $\bm{T}^e$.
To align domain knowledge with the SLM, we construct a domain-specific vocabulary $\bm{V}^k$ by filtering standard language vocabulary $\bm{V}$ based on structured medical knowledge $\bm{K}$:
\begin{equation}
\bm{V}^k = \{v_1, \dots, v_n \mid v_1 \in \bm{K}, \bm{K} \subseteq \bm{V} \}. 
\end{equation}
A cross-attention mechanism is then applied to integrate knowledge-driven representations into the model. The lab test embeddings ($\bm{T}^e$) serve as the Query, while domain knowledge tokens ($\bm{V}^k$) act as the Key and Value:
\begin{equation}
\begin{aligned}
&\bm{H} =  f_\phi(\bm{T},\bm{V}^k),\\
&=SoftMax\Big(\frac{(\bm{T}^e\bm{W}^q)({\bm{V}^k}^{\top}\bm{W}^k)}{\sqrt{d}}\Big)(\bm{V}^k\bm{W}^v),\\
\end{aligned}
\end{equation}
where $d$ is the hidden dimension of the SLM, and $\bm{W}^q$, $\bm{W}^k$, $\bm{W}^v$ are learnable attention weight matrices, $f$ indicates the encoding function of the TSE and attention, and $\phi$ represents the trainable parameters of $f$. The resulting knowledge-enhanced embeddings $\bm{H}$ are then fed into the frozen distilled SLM to refine its reasoning and generate lab test-based rationales ($\bm{R}^t$) and diagnosis ($\bm{D}$).

The model is trained using the following objective function:
\begin{equation}
\mathcal{L}_{lab}(\phi) =\mathbb{E}[-\log P_{SLM_{\theta}}(\bm{D}, \bm{R}^t \mid \bm{H})].
\end{equation}
 This phase ensures that the SLM can naturally interpret lab tests while effectively leveraging medical knowledge to enhance its reasoning capabilities.

\subsubsection{Phase 3: Full Multimodal Rationale Distillation}
In the final phase, ClinRaGen is trained to generate full multimodal clinical rationales by integrating medical notes, lab tests, and structured domain knowledge. To ensure effective multimodal reasoning, lab test $\bm{T}$ is formatted as prefix prompts~\cite{niu2024ehr}, allowing the SLM to seamlessly incorporate it with textual EHRs.

During this stage, the model is optimized to generate both medical note-based rationales ($\bm{R}^m$) and lab test-based rationales ($\bm{R}^t$), ensuring that all available information contributes to clinically coherent and interpretable decision-making. The multimodal rationale distillation objective is formulated as follows:
\begin{equation}
\begin{aligned}   
\mathcal{L}_{mm}(\theta, \phi) = &\mathbb{E}[-\log P_{SLM_{\theta}}(\bm{D}, \bm{R}^m, \bm{R}^t  \mid \\
&\bm{M},f_{\phi}(\bm{T},\bm{V}^k))].
\end{aligned}
\end{equation}
The fine-tuning of all ClinRaGen components, ensuring that multimodal EHRs are effectively integrated, enhances diagnostic accuracy and produces modality-consistent rationales.

\section{Experiments}

\begin{table*}[t]
\setlength\tabcolsep{3.5pt}
\renewcommand\arraystretch{1}
\centering
\small

\begin{tabular}{lrcccccccc}
\toprule[2pt]

\multirow{2}{*}{\textbf{Models}} &\multirow{2}{*}{\textbf{Size}} &\multicolumn{2}{c}{\textbf{Modality}} &\multicolumn{3}{c}{\textbf{Micro}} &\multicolumn{3}{c}{\textbf{Macro}}   \\ 
                            \cmidrule{3-10} 
                           && \multicolumn{1}{c}{\textbf{Lab}} & \textbf{Note}  &\multirow{1}{*}{\textbf{Precision}} & \multirow{1}{*}{\textbf{Recall}} & \multirow{1}{*}{\textbf{F1}}&\multirow{1}{*}{\textbf{Precision}} & \multirow{1}{*}{\textbf{Recall}} & \multirow{1}{*}{\textbf{F1}} \\ 
\toprule[1pt]
\multicolumn{10}{c}{\textbf{ MIMIC-III}}  \\ 
\hline
Flan-T5 &  60M   &  & \checkmark & 0.5812$_{(0.11)}$  & 0.6623$_{(0.07)}$ & 0.6203$_{(0.05)}$ &  0.5656$_{(0.10)}$ & 0.6247$_{(0.08)}$   & 0.5887$_{(0.07)}$  \\
\hline
PROMPTEHR  & 75.2M& &  \checkmark &  0.5929$_{(0.11)}$  & 0.6553$_{(0.07)}$ & 0.6224$_{(0.02)}$ & 0.5744$_{(0.10)}$ & 0.6287$_{(0.06)}$   & 0.5910$_{(0.03)}$  \\
\hline
LLaMA-ft  & 7B & \checkmark & \checkmark &  0.6142$_{(0.21)}$  & 0.6598$_{(0.15)}$  & 0.6364$_{(0.04)}$  & 0.6108$_{(0.15)}$  &  0.6164$_{(0.13)}$  & 0.6055$_{(0.04)}$    \\
\hline
FROZEN &265M & \checkmark  & \checkmark &  0.6102$_{(0.18)}$ & 0.6401$_{(0.16)}$ & 0.6231$_{(0.03)}$  &  0.5976$_{(0.16)}$ & 0.6001$_{(0.17)}$ & 
0.5915$_{(0.03)}$ \\
\hline
EHR-KnowGen &  77M&\checkmark  & \checkmark & 0.6001$_{(0.03)}$ & 0.6551$_{(0.02)}$ & 0.6262$_{(0.01)}$ & 0.5834$_{(0.04}$ &  0.6181$_{(0.03)}$ & 0.5944$_{(0.01)}$  \\
\hline
Clinical CoT   & && &&&&&&\\
-w/o TSE & 60M&& \checkmark & 0.6115$_{(0.03)}$ & 0.6402$_{(0.04)}$ &  0.6311$_{(0.03)}$ & 0.6024$_{(0.04)}$ & 0.5989$_{(0.06)}$ & 0.5969$_{(0.03)}$ \\
-w/ TSE &85M&\checkmark& \checkmark & 0.5967$_{(0.05)}$ & 0.6607$_{(0.06)}$ &  0.6328$_{(0.03)}$ & 0.5924$_{(0.06)}$ & 0.6092$_{(0.07)}$ & 0.5975$_{(0.05)}$ \\
\hline
LLM Zero-shot & && &&&&&&\\

-LLaMA & 7B& \checkmark  & \checkmark & 0.1227$_{(0.08)}$ & 0.0421$_{(0.06)}$ &  0.0627$_{(0.06)}$ & 0.0392$_{(0.06)}$  & 0.0622$_{(0.06)}$  & 0.0438$_{(0.05)}$   \\
-ChatGPT &  175B& \checkmark  & \checkmark &  0.4474$_{(0.07)}$  & 0.1405$_{(0.05)}$ & 0.2139$_{(0.05)}$ & 0.4883$_{(0.08)}$ & 0.1872$_{(0.05)}$  &  0.2188$_{(0.04)}$   \\
\hline
\rowcolor{gray!20} ClinRaGen & 87M& \checkmark  & \checkmark & 0.6104$_{(0.02)}$  & 0.6751$_{(0.02)}$  & \underline{0.6410}$_{(0.01)}$ &  0.5991$_{(0.03)}$  &  0.6311$_{(0.04)}$  &  \underline{0.6113}$_{(0.02)}$   \\
\rowcolor{gray!20} ClinRaGen* & 793M& \checkmark  & \checkmark & 0.6047$_{(0.03)}$  & 0.6875$_{(0.03)}$  & \textbf{0.6501}$_{(0.02)}$ &  0.5943$_{(0.04)}$  &  0.6531$_{(0.03)}$  &  \textbf{0.6196}$_{(0.03)}$   \\
\hline

\multicolumn{10}{c}{\textbf{ MIMIC-IV}}  \\ 
\hline

Flan-T5  &  60M & & \checkmark &  0.6624$_{(0.05)}$ & 0.6953$_{(0.02)}$ & 0.6792$_{(0.04)}$ & 0.6428$_{(0.06)}$ &  0.6601$_{(0.05)}$ & 0.6479$_{(0.04)}$  \\
\hline
PROMPTEHR& 75.2M & & \checkmark &  0.6524$_{(0.07)}$ & 0.7031$_{(0.06)}$ & 0.6802$_{(0.02)}$ & 0.6353$_{(0.05)}$ & 0.6702$_{(0.07)}$ & 0.6501$_{(0.03)}$  \\
\hline
LLaMA-ft   &  7B & \checkmark  & \checkmark &  0.6854$_{(0.11)}$ & 0.6954$_{(0.07)}$ & 0.6929$_{(0.03)}$ &   0.6753$_{(0.09)}$ & 0.6624$_{(0.11)}$ & 0.6621$_{(0.06)}$   \\
\hline
FROZEN &265M & \checkmark  & \checkmark & 0.6781$_{(0.08)}$ & 0.6908$_{(0.09)}$ & 0.6842$_{(0.01)}$ & 0.6627$_{(0.10)}$ & 0.6521$_{(0.10)}$ & 0.6530$_{(0.02)}$  \\
\hline
EHR-KnowGen  & 77M& \checkmark  & \checkmark & 0.6580$_{(0.06)}$  & 0.7085$_{(0.05)}$ &  0.6816$_{(0.02)}$ & 0.6382$_{(0.05)}$ & 0.6724$_{(0.06)}$ & 0.6511$_{(0.02)}$   \\
\hline
Clinical CoT  & && &&&&&&\\

-w/o TSE &60M &  & \checkmark & 0.6751$_{(0.05)}$ &  0.7069$_{(0.03)}$ &  0.6905$_{(0.03)}$ & 0.6607$_{(0.04)}$ & 0.6796$_{(0.06)}$ &  0.6612$_{(0.02)}$\\
-w/ TSE  &85M & \checkmark & \checkmark & 0.7011$_{(0.04)}$ &  0.6808$_{(0.06)}$ &  0.6917$_{(0.04)}$ & 0.6971$_{(0.05)}$ & 0.6354$_{(0.03)}$ &  0.6577$_{(0.03)}$\\
\hline

LLM Zero-shot & && &&&&&&\\

-LLaMA & 7B& \checkmark  & \checkmark & 0.1357$_{(0.11)}$ & 0.0997$_{(0.07)}$ &  0.1150$_{(0.06)}$ &  0.0435$_{(0.09)}$  &  0.1466$_{(0.07)}$ &   0.0619$_{(0.05)}$  \\
-ChatGPT &  175B& \checkmark  & \checkmark & 0.4536$_{(0.07)}$ & 0.1458$_{(0.05)}$  & 0.2207$_{(0.04)}$  &   0.4532$_{(0.06)}$   &   0.1831$_{(0.06)}$  &   0.2147$_{(0.05)}$   \\
\hline

\rowcolor{gray!20}ClinRaGen&87M& \checkmark  & \checkmark &  0.7009$_{(0.01)}$ & 0.6963$_{(0.02)}$ & \underline{0.6989}$_{(0.01)}$ &  0.6868$_{(0.03)}$ &  0.6603$_{(0.01)}$ &   \underline{0.6685}$_{(0.02)}$ \\
\rowcolor{gray!20}ClinRaGen*  &793M& \checkmark  & \checkmark &  0.6848$_{(0.04)}$ & 0.7429$_{(0.02)}$ &  \textbf{0.7127}$_{(0.02)}$ &  0.6779$_{(0.02)}$ &  0.7087$_{(0.01)}$ &  \textbf{0.6893}$_{(0.01)}$ \\
\hline
\toprule[2pt]
\end{tabular}
\caption{ The performance of comparative methods in the disease diagnosis tasks on MIMIC-III and MIMIC-IV. The best results are
highlighted in bold, and the second-best results are marked with an underline. }
\label{table1}
\end{table*}

\subsection{Experimental Settings}
\textbf{Dataset:}
We evaluate ClinRaGen on two public EHR datasets: MIMIC-III \cite{johnson2016mimic} (28,456 EHRs include medical notes and time series lab tests) and MIMIC-IV \cite{johnson2023mimic} (28,900 EHRs). Both datasets use benchmark tools \cite{Harutyunyan2019} for time series processing, with missing values filled by the nearest available data. We target 25 disease phenotypes and follow a 4:1 training-to-testing split \cite{Harutyunyan2019}. Our model is available at github\footnote{\href{https://github.com/Healthcare-Data-Mining-Laboratory/ClinRaGen}{https://github.com/Healthcare-Data-Mining-Laboratory/ClinRaGen}}.

\textbf{Baseline Methods:}
To evaluate the effectiveness of ClinRaGen for disease diagnosis generation, we compared it with following baselines: Flan-T5 \cite{chung2024scaling}, PROMPTEHR \cite{wang2022promptehr}, FROZEN \cite{tsimpoukelli2021multimodal}, EHR-KnowGen \cite{niu2024ehr}, Clinical CoT (with/without TSE) \cite{kwon2024large}, and LLM-based models LLaMA-7B \cite{touvron2023llama} (zero-shot and fine-tuning) and ChatGPT \cite{open2023chatgpt} (zero-shot). Baseline and implementation details are provided in Appendices A.3 and A.4. For a fair comparison, all baselines (except LLaMA) use Flan-T5-Small as the backbone; our model employs Flan-T5-Small (ClinRaGen) and  Flan-T5-Large (ClinRaGen*) to evaluate the effect of varying scales. ChatGPT (GPT-3.5-turbo) serves as our teacher LLM. Results are averaged over five runs, with statistical significance determined at p < 0.05 by t-test.

\begin{table}[htbp]
\setlength\tabcolsep{10pt}
\renewcommand\arraystretch{1}
\centering
\small

\begin{tabular}{lcc}
\toprule[1pt]
\multirow{1}{*}{\textbf{Models}} 
                           &\textbf{Micro F1} & \textbf{Macro F1}\\ 
                          \toprule[1pt]
\multicolumn{3}{c}{\textbf{MIMIC-III }} \\
  \hline  
  
ClinRaGen     &  \textbf{0.6410}  &   \textbf{0.6113}  \\
 \cline{1-3}
 \textit{w/o} LAB\&KNOW     &  0.6323 & 0.6021  \\
 \textit{w/o} KNOW    &  0.6349 & 0.6042 \\
 \textit{w/o} REASONING     &  0.6255 & 0.5915  \\
  \hline  
\multicolumn{3}{c}{\textbf{MIMIC-IV}}\\
  \hline  
ClinRaGen     &  \textbf{0.6989} &  \textbf{0.6685}  \\
 \cline{1-3}
 \textit{w/o} LAB\&KNOW    &  0.6925 &  0.6643  \\
 \textit{w/o} KNOW   &  0.6936 &  0.6644  \\
 \textit{w/o} REASONING   &  0.6828 & 0.6541  \\
\toprule[1pt]
\end{tabular}
\caption{Ablation studies on disease diagnosis. }
\label{ablation}
\end{table}

\begin{figure*}[t]
\centering
\centerline{\includegraphics[scale=0.36 ]{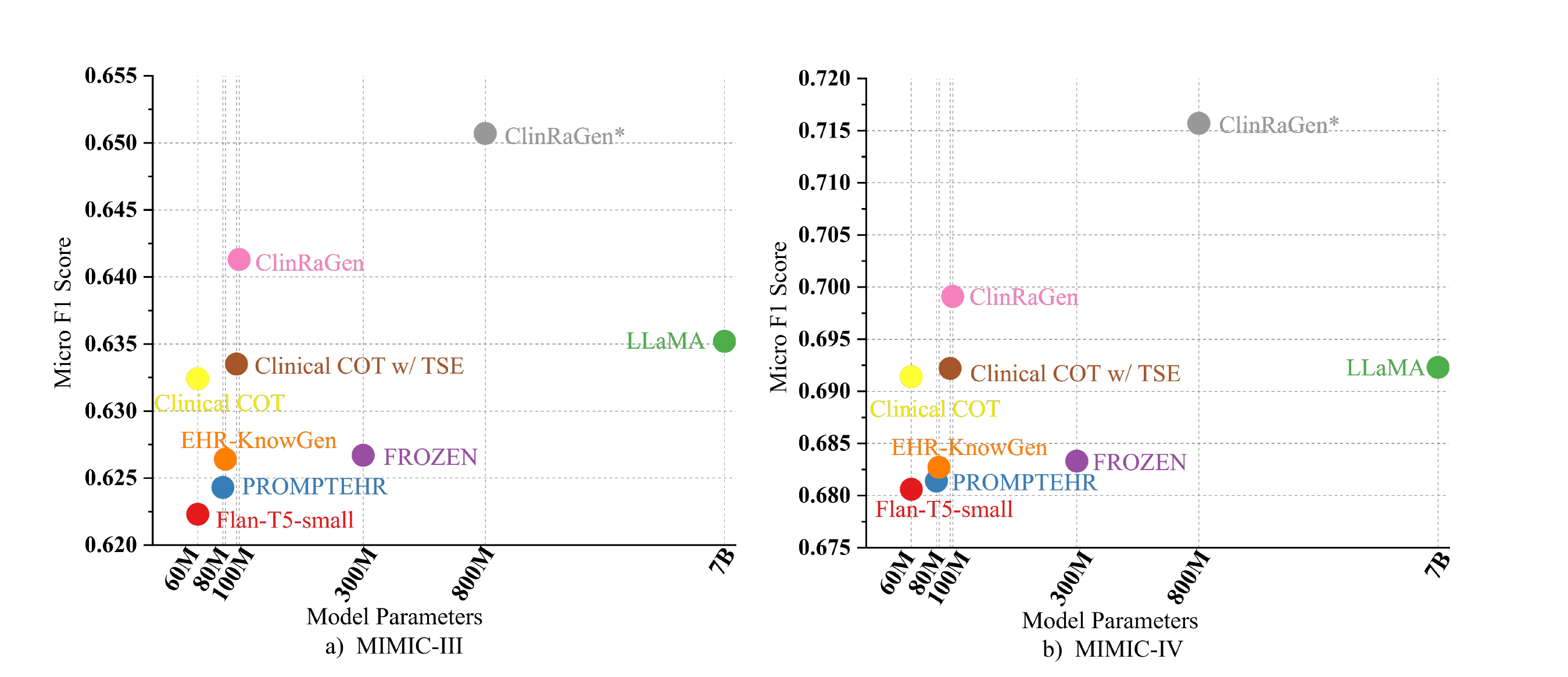}}
\caption{Model Parameter Counts and Micro F1 Scores}
\label{model_parameters}
\end{figure*}

\subsection{Disease Diagnosis Performance}
\textbf{Comparison with Baselines:}
We evaluate disease diagnosis using micro and macro precision, recall, and F1 scores. Table~\ref{table1} shows that multimodal models outperform single-modality models, confirming the value of lab test results. Clinical CoT surpasses other baseline models, highlighting the rationale for distillation's effectiveness. ClinRaGen (80M) achieves the best performance, with an average F1 score improvement of over 1.1\% across all baselines, even outperforming LLaMA-7B-ft. Furthermore, ClinRaGen* (793M) improves by over 1.5\%, significantly exceeding other baselines. The weak performance of zero-shot LLMs confirms the absence of data leakage. These results demonstrate ClinRaGen's ability to match or surpass LLMs in clinical tasks through multimodal rationale distillation and the knowledge-augmented attention mechanism. Appendix A.6 provides further experimental details, revealing a substantial improvement of over 3\% in F1 scores achieved by our method across different SLMs and through distillation from varied LLMs.

\textbf{Ablation Study:}  
We assess the impact of key components in ClinRaGen: (1) \textit{w/o} LAB\&KNOW removes lab tests and knowledge input, (2) \textit{w/o} KNOW replaces the knowledge-based vocabulary with a standard one, and (3) \textit{w/o} REASONING excludes rationale distillation while maintaining model structure. Table~\ref{ablation} shows that \textit{w/o} REASONING performs worst, highlighting the importance of rationale distillation. The drop in F1 scores for \textit{w/o} LAB\&KNOW confirms the value of multimodal integration, while \textit{w/o} KNOW shows the KA module’s contribution to diagnostic accuracy.

\textbf{Model Efficiency:}  
We evaluate ClinRaGen’s efficiency by comparing model parameters, micro F1 scores (Figure~\ref{model_parameters}), and training times (Table~\ref{time_evaluation}). ClinRaGen (80M) achieves superior diagnostic performance with 80× fewer parameters and less than half the training time of LLaMA-7B. These results highlight the effectiveness of our sequential multimodal distillation paradigm and Knowledge-augmented attention mechanism in enabling efficient and accurate clinical reasoning.

\begin{table}[htbp]
\setlength\tabcolsep{4pt}
\renewcommand\arraystretch{0.95}
\centering
\small

\begin{tabular}{lc}
\toprule[1pt]
{\textbf{Models}}  & Time Cost (Seconds)\\

\hline
Knowledge Retrieval  &12,636\\
LLM-Guided Rationale Generation	&604,715\\
LlaMA – 7B Tuning	&259,113\\
ClinRaGen – 84M Tuning	&94,623\\

\toprule[1pt]
\end{tabular}
\caption{Time Cost Evaluation. }
\label{time_evaluation}
\end{table}
\subsection{Rationale Evaluation with LLM-as-a-Judge and Human }  

\textbf{Evaluation Methods:} To assess the quality of generated multimodal rationales and maximize the potential of SLMs, we evaluate ClinRaGen (80M) using five evaluation criteria: \emph{Correctness}, \emph{Readability}, \emph{Soundness}, \emph{Consistency}, and \emph{Persuasiveness}, as informed by clinicians and prior research \cite{lin2024towards,kwon2024large}. These criteria are scored on a Likert scale ranging from 1 to 5 (details of the criteria are provided in Appendix A.5). We conduct both LLM-based evaluations (using LLM-as-a-Judge \cite{gu2024survey}) and human evaluations. For LLM comparisons, we use Mistral-7B, LLaMA2-7B, and LLaMA3-8B with five-shot prompting. Distilled rationales from ChatGPT serve as ground truth (GT). Comparative LLMs receive time series anomalies and medical notes, while ClinRaGen directly processes numerical lab test and medical notes.  
Following \citet{lin2024towards,chiang2023can}, we use GPT-4 to evaluate 1000 randomly selected samples.  For human assessment, 15 professional postgraduates rate 100 samples, achieving moderate intra-class (0.637) and inter-class (0.608) agreement, indicating reasonable consistency despite the task's subjectivity.

\begin{figure}[t]
\centering
\centerline{\includegraphics[scale=0.37 ]{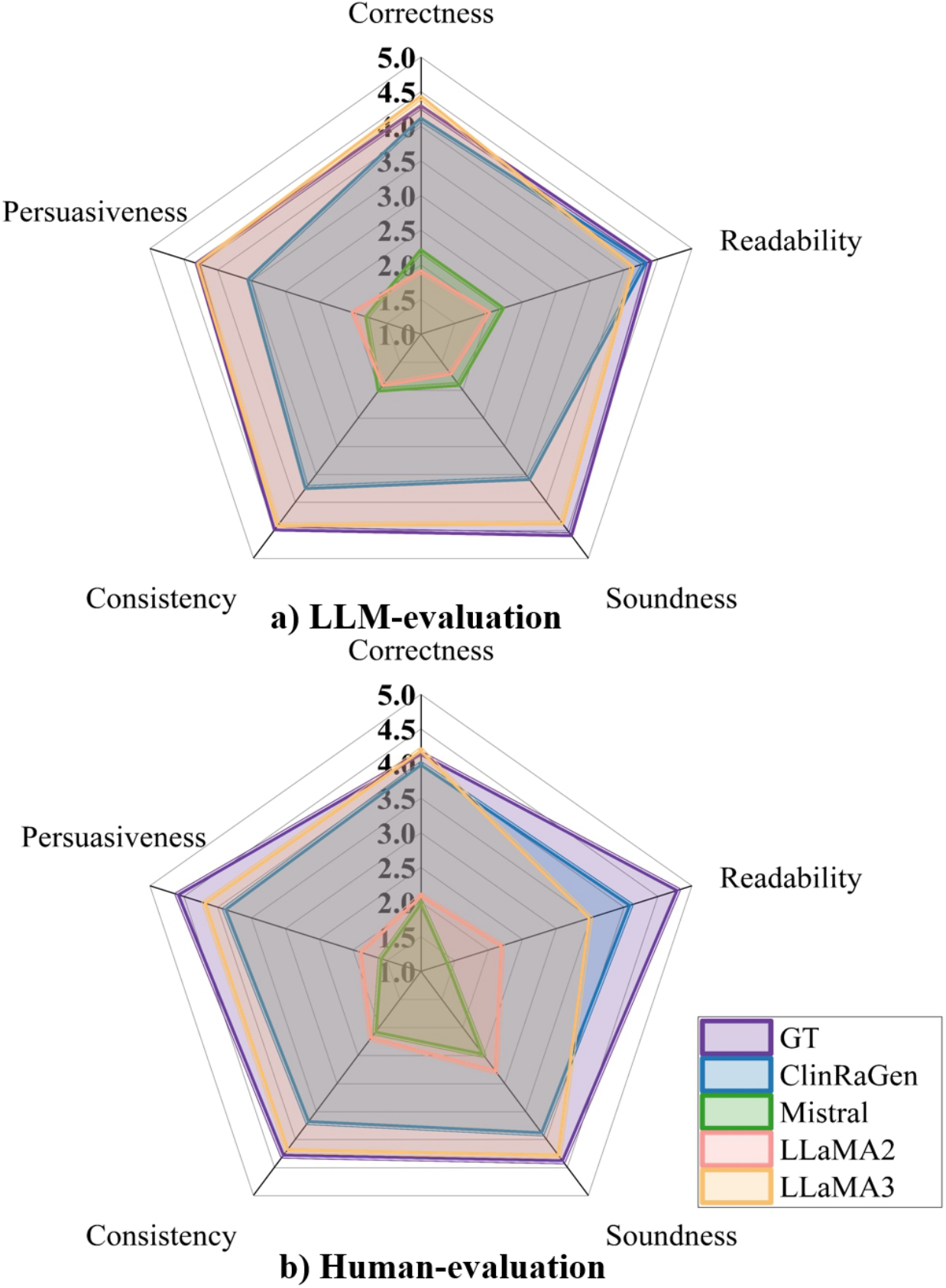}}
\caption{Rationale evaluation with LLMs and Human }
\label{rationale_eval_fig}
\end{figure}

\textbf{Evaluation Results:}  
Figures~\ref{rationale_eval_fig}(a) and (b) show GPT-4 and human evaluations across five criteria, with closely aligned results. LLaMA3 performs best, benefiting from its large scale and pre-training. ClinRaGen ranks second, matching LLaMA3 in readability and correctness while surpassing LLaMA2 and Mistral, which often generate incoherent rationales. Unlike other LLMs relying on anomaly captions, ClinRaGen achieves the second-highest consistency score, demonstrating the knowledge-augmented attention mechanism’s effectiveness in consistent multimodal reasoning. ClinRaGen also outperforms LLaMA2 and Mistral in soundness and persuasiveness, further underscoring our method's effectiveness.

\subsection{Rationale Evaluation with BLUE and BERTScore.} \label{Rationale Evaluation with BLUE and BERTScore}

In addition to the criteria defined for evaluating rationale performance, Table~\ref{rationale_eval} presents the performance of our model, ClinRaGen, alongside various baselines, using both BLEU \cite{papineni2002bleu} and BERTScore \cite{zhang2019bertscore} on the MIMIC-III and MIMIC-IV datasets.  The results show that ClinRaGen outperformed all other models in both metrics across the datasets. The latest open-source LLM, LLaMA3, ranked second, while Mistral exhibited the poorest performance. These results are consistent with those from LLM and human evaluations.

\begin{figure}[htbp]
\centering
\centerline{\includegraphics[scale=0.42]{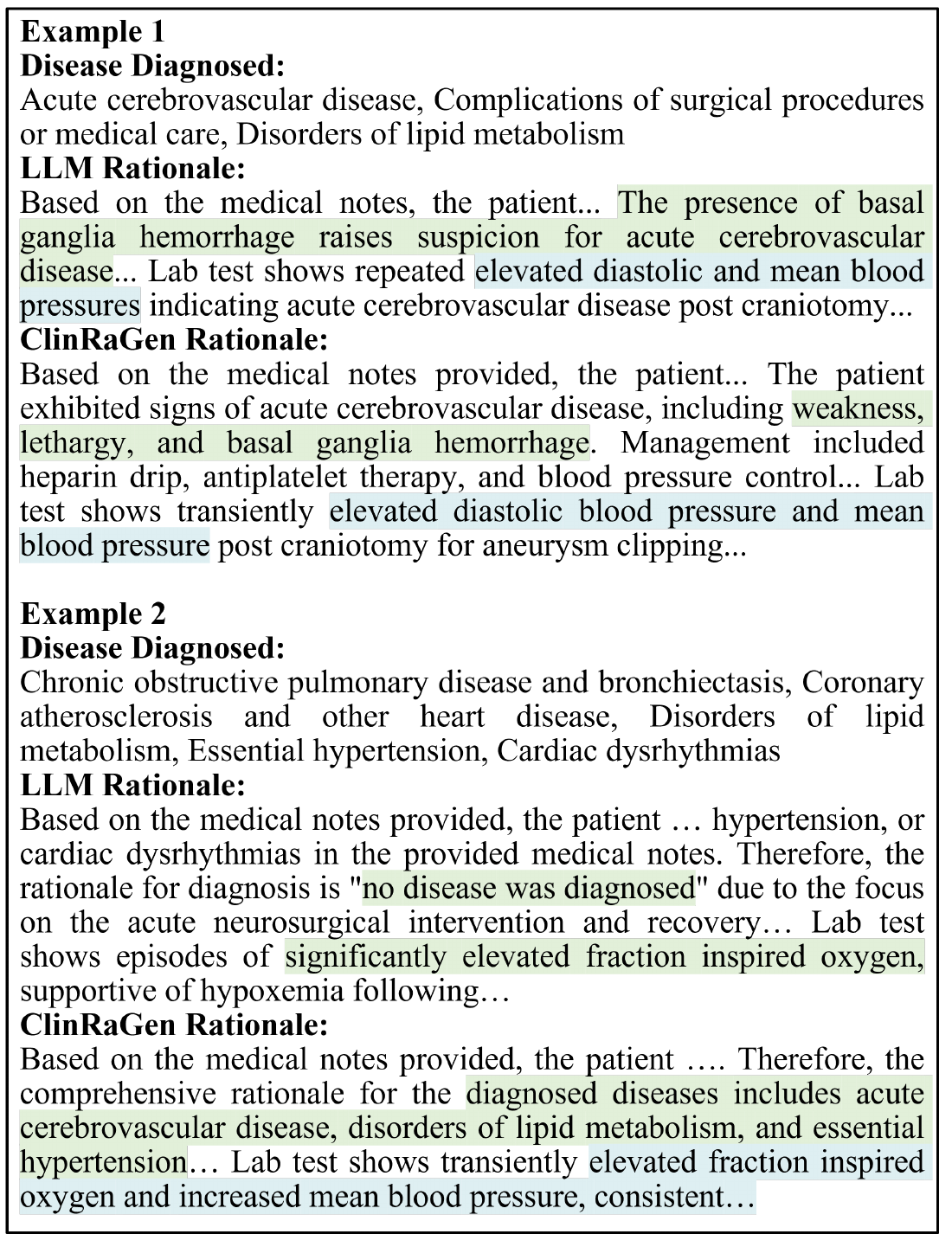}}
\caption{Case studies on disease diagnosis and clinical rationale generation compared with teacher LLM.}
\label{casestudy}
\end{figure}

\textbf{Case Studies:} 
As illustrated in Figure~\ref{casestudy}, our model ClinRaGen can produce both medical note-based rationales (e.g., ``\emph{Based on the medical notes...}") and lab test-based rationales (e.g., ``\emph{Lab test shows...}"), akin to the outputs of teacher LLM. For medical note-based rationale generation, ClinRaGen effectively extracts key medical terms essential for disease diagnosis (highlighted in green). Additionally, for lab test-based rationales, our model accurately identifies abnormal lab test features (highlighted in blue), demonstrating its capability to understand numerical time series lab test data effectively. These results indicate that ClinRaGen competently produces clinically relevant multimodal rationales to support disease diagnosis.
\begin{table}[t]
\setlength\tabcolsep{4pt}
\renewcommand\arraystretch{1}
\centering
\small
\begin{tabular}{lcccc}
\toprule[1pt]
\multirow{2}{*}{\textbf{Models}}  & \multicolumn{2}{c}{\textbf{MIMIC-III }} & \multicolumn{2}{c}{\textbf{MIMIC-IV}}\\

            \cline{2-5}
                           &\textbf{BLEU} & \textbf{BERTScore}&\textbf{BLEU} & \textbf{BERTScore}\\ 
                          \toprule[1pt]

Mistral       &  0.0163 & 0.7348 & 0.0532 & 0.7625  \\
LLaMA2   &  0.1441 & 0.8714 & 0.2357 & 0.8808  \\
LLaMA3      &  0.1641 & 0.8804 & 0.2568 &  0.8919 \\
ClinRaGen     &  \textbf{0.2689}  &   \textbf{0.8972} &  \textbf{0.2963} &  \textbf{0.9044}  \\
\toprule[1pt]
\end{tabular}
\caption{Rationale evaluation with BLEU and BERTScore. }
\label{rationale_eval}
\end{table}

\textbf{Further Discussions:}  
LLMs may introduce bias into distilled clinical rationales. To evaluate ClinRaGen’s correctness, we assess the relevance of key medical terms to diagnosed diseases. In one case, our model identifies \emph{weakness}, \emph{lethargy}, and \emph{basal ganglia hemorrhage} as evidence for \emph{acute cerebrovascular}, while teacher LLM  captures only \emph{basal ganglia hemorrhage}, missing relevant symptoms \cite{Unnithan2023-vp}. In another case, while the teacher LLM reports \emph{no disease}, ClinRaGen correctly identifies conditions like \emph{disorders of lipid metabolism} and \emph{essential hypertension}. These results highlight ClinRaGen’s ability to mitigate LLM biases by capturing time-series variations and integrating structured knowledge.

\section{Conclusion and Future Work}  

We present ClinRaGen, a knowledge-augmented framework that enhances SLMs with LLM-derived reasoning and structured medical knowledge for disease diagnosis and multimodal rationale generation. It introduces a knowledge-augmented attention module that unifies time-series and textual EHRs within a shared encoding space while incorporating domain knowledge for reliable multimodal rationale generation. Additionally, it presents a sequential multimodal distillation paradigm for transferring the reasoning capabilities of LLMs to SLMs. The extensive evaluations performed on two real-world medical datasets, MIMIC-III and MIMIC-IV, which included both quantitative and qualitative analyses, demonstrate that ClinRaGen allows SLMs to achieve performance comparable to LLMs in disease diagnosis and the generation of multimodal rationales. Furthermore, it can mitigate the bias inherent in distilled rationales from LLMs. This work effectively bridges the performance gap between LLMs and SLMs in clinical tasks.
Future research will extend ClinRaGen to a broader range of SLM architectures, LLMs,  datasets, and medical applications.  

\section*{Limitations}  

While ClinRaGen effectively enhances multimodal clinical reasoning, certain limitations remain:
\begin{itemize}
\item First, although knowledge-augmented rationale distillation transfers reasoning capabilities from LLMs to SLMs, some potential biases in LLM-generated rationales may still persist. 
\item  Second, the effectiveness of the knowledge-augmented attention module depends on the quality and coverage of external knowledge sources. 
\item Lastly, ClinRaGen is evaluated on structured EHR datasets, and its applicability to unstructured clinical text or other medical modalities requires further exploration.
\end{itemize}
Future work will refine knowledge integration, enhance more effective bias mitigation strategies, and extend evaluations to diverse clinical settings.

\section*{Ethics Statement}

\paragraph{Data Privacy:} The datasets utilized in our research are publicly accessible and feature de-identified patient data; accessing these datasets still requires passing the CITI Exam\footnote{https://about.citiprogram.org/} and downloading from Physionet\footnote{https://physionet.org/}. In addition, this study used the Azure OpenAI service and completed the "opting out of the review process" agreement.

\section*{Acknowledgments}
This work is supported by the National Natural Science Foundation of China (No.62432006, No.62276159), Tencent Rhino-Bird Focused Research Program (Value-aligned Credible Large Language Model) and RMGS project (Artificial Intelligence and Big Data Analytics for Social Good).

\bibliography{main}
\bibliographystyle{acl_natbib}

\appendix

\section{Appendix}
\label{sec:appendix}

\begin{table*}[t]
\setlength\tabcolsep{1.7pt}
\centering
\begin{tabular}{l}
\toprule[1pt]
Condition: If the lab test value is not an abnormal value:  \\
Prompt: \{Lab features\} is normal all the time.\\
\hline
Condition: If the lab test value is an abnormal value higher than the standard: \\
Prompt: \{Lab features\} is higher than normal \{number of times\} times.\\
\hline
Condition: If the lab test value is an abnormal value lower than the standard: \\
Prompt: \{Lab features\} is lower than normal \{number of times\} times.\\
\hline
Condition: If the lab test value is abnormal, it includes both higher and lower than the standard value: \\
Prompt: \{Lab features\} is higher than normal \{number of times\} times and lower than normal \\
\{number of times\} times.\\
\hline

\toprule[1pt]
\end{tabular}
\caption{Lab test anomaly caption template. }
\label{anomaly}
\end{table*}

\subsection{Lab Test Anomaly Caption}
To caption lab test results into textual descriptions, we initially employ the Inter Quartile Range (IQR) anomaly detection method \cite{vinutha2018detection} to identify anomalous lab test features. Subsequently, we craft multiple text templates to caption these anomalies. These templates are delineated in Table~\ref{anomaly}.

\subsection{Prompts for Multimodal Rationale Generation Via LLMs} 
The overall procedure for teacher LLMs generating clinical rationale is illustrated in Figure~\ref{model1}. The specific medical note-based rationale prompt and lab test-based rationale prompt are detailed as follows. 
\textbf{Medical note-based rationale prompt} for LLMs,

``\emph{Below is an instruction that describes examples of generating the rationale of disease diagnosis; please refer to the examples style to generate the Output from the Input:}

\emph{\#\#\# Instruction:}

\emph{There are some examples please to refer: }

\emph{Example 1, Example 2, Example ... }

\emph{\#\#\# Input:}

\emph{\#\#\# Medical note: [$\bm{M}$]}

\emph{\#\#\# Diagnosed diseases: [$\bm{D}$]}

\emph{Please review the patient's medical records. Adhere to the provided format to craft a succinct 100-word rationale for diagnosing these conditions (Start with "Based on the medical notes..."). If the diagnosis indicates "no disease was diagnosed," the rationale must state "no disease was diagnosed." Otherwise, provide a comprehensive rationale for the diagnosis.}

\emph{\#\#\# Response:}

\emph{\#\#\# Output:}

\emph{\#\#\# Medical note-based Rationale: [$\bm{R}^n$]}

\textbf{Lab test-based rationale prompt} for LLMs, we denote the lab test anomalies as $\bm{T}^*$:

``\emph{Below is an instruction that describes examples of generating the rationale of disease diagnosis, please refer to the examples style to generate the Output from the Input:}

\emph{\#\#\# Instruction:}

\emph{There are some examples please to refer: }

\emph{Example 1, Example 2, Example ... }

\emph{\#\#\# Input:}

\emph{\#\#\# Medical note: [$\bm{M}$]}

\emph{\#\#\# Descriptions of lab test abnormalities: [$\bm{T}^*$]}

\emph{\#\#\# Diagnosed diseases: [$\bm{D}$]}

\emph{\#\#\# Medical note-based rationale: [$\bm{R}^n$]}

\emph{Please review the patient's medical notes, laboratory test anomaly results, and existing rationales in the medical record. Construct a concise, one-sentence rationale, limited to max 50 words, that accurately describes a diagnosed condition based on descriptions of laboratory test abnormalities (Start with "Lab test shows..."). Pay close attention to potential inaccuracies in the lab descriptions.
}

\emph{\#\#\# Response:}

\emph{\#\#\# Output:}

\emph{\#\#\# Lab test-based rationale: [$\bm{R}^t$]}

\begin{table*}[t]
\centering
\small
\setlength{\tabcolsep}{15pt}
\renewcommand{\arraystretch}{1.2}
\begin{tabular}{lccccc}
\toprule
\multirow{2}{*}{\textbf{Base SLMs}} & \multirow{2}{*}{\textbf{Method}} &  \multicolumn{2}{c}{\textbf{Distill From ChatGPT}} & \multicolumn{2}{c}{\textbf{Distill From DeepSeek}} \\
\cmidrule(r){3-6}
 & &\textbf{Micro F1} & \textbf{Macro F1} & \textbf{Micro F1} & \textbf{Macro F1} \\
\midrule
\multicolumn{6}{c}{\textbf{MIMIC-III}} \\
\midrule
\multirow{2}{*}{Flan-T5-0.1B }           &  Clinical-COT & 0.6323        & 0.6021        & 0.6338 & 0.6003 \\  
& ClinRaGen   & \textbf{0.6410} & \textbf{0.6113} &\textbf{0.6447} & \textbf{0.6121} \\
\midrule

\multirow{2}{*}{Flan-T5-0.7B  }            &  Clinical-COT   &  0.6398     &  0.6059     & 0.6403 & 0.6072 \\  
   & ClinRaGen & \textbf{0.6501} & \textbf{0.6196} &  \textbf{0.6509} &  \textbf{0.6222} \\
\midrule

\multirow{2}{*}{OPT-0.1B }               & Clinical-COT & 0.4573        & 0.4099        &  0.4571 & 0.4092 \\ 
   & ClinRaGen & \textbf{0.4929} & \textbf{0.4514}  & \textbf{0.5009} & \textbf{0.4555} \\
\midrule
\multirow{2}{*}{QWEN-2.5-0.5B }               &  Clinical-COT &   0.5921     &   0.5400  & 0.5950 & 0.5566 \\  
  & ClinRaGen & \textbf{0.6207} & \textbf{0.5688}  & \textbf{0.6219} & \textbf{0.5775}  \\
\midrule
\multicolumn{6}{c}{\textbf{MIMIC-IV}} \\
\midrule
\multirow{2}{*}{Flan-T5-0.1B  }             &   Clinical-COT&    0.6925    &   0.6643      &0.6933& 0.6647\\  
  & ClinRaGen & \textbf{0.6989} & \textbf{0.6685} & \textbf{0.6994} & \textbf{0.6712} \\
\midrule

\multirow{2}{*}{Flan-T5-0.7B }             &   Clinical-COT &   0.7055     &    0.6801    &  0.7062 & 0.6803\\  
   & ClinRaGen& \textbf{0.7127} & \textbf{0.6893}& \textbf{0.7129} &  \textbf{0.6896} \\ 
\midrule

\multirow{2}{*}{OPT-0.1B }               &  Clinical-COT &  0.5067   &  0.4547    & 0.5063 & 0.4533\\
   & ClinRaGen & \textbf{0.5284} & \textbf{0.4810} & \textbf{0.5315} & \textbf{0.4884} \\
\midrule
\multirow{2}{*}{QWEN-2.5-0.5B }             &  Clinical-COT&   0.6114 & 0.5882 & 0.6095 & 0.5351\\  
   & ClinRaGen &   \textbf{0.6332}    &   \textbf{0.6105}     & \textbf{0.6406} & \textbf{0.5583} \\
\bottomrule

\end{tabular}
\caption{Ablation studies on different SLMs and LLMs choice.}
\label{table_appendix}
\end{table*}
\subsection{Baseline Details}\label{Baseline Details}

\begin{itemize}

\item \textbf{Flan-T5}: Flan-T5 is introduced in the scaling instruction-fine-tuning method for language models \cite{chung2024scaling}. It is trained on comprehensive datasets designed for tasks like summarization, question answering, and reasoning, enhancing its chain-of-thought capabilities.

\item \textbf{PROMPTEHR}: PROMPTEHR \cite{wang2022promptehr} innovates generative modelling for EHRs through conditional prompt learning; in this experiment, we focus on applying it, particularly on disease diagnosis.

\item \textbf{LLaMA}: The LLaMA-7B model \cite{touvron2023llama}, a prominent large language model, employs Reinforcement Learning with Human Feedback (RLHF) and instructional tuning, showcasing its adaptability across diverse NLP tasks. This study applied both zero-shot and fine-tuning for disease diagnosis.

\item \textbf{FROZEN}: The FROZEN framework \cite{tsimpoukelli2021multimodal} stands out in multimodal vision-language modeling for few-shot learning. Here, it's tailored to disease diagnosis, analyzing both lab test results and medical notes.

\item \textbf{EHR-KnowGen}: As a leading model in EHR multimodal learning, EHR-KnowGen \cite{niu2024ehr} specializes in generating disease diagnoses. This study excludes external knowledge to maintain a balanced evaluation.

\item \textbf{Clinical CoT}: Clinical CoT \cite{kwon2024large} integrates clinical reasoning into a diagnostic framework for EHRs using prompt-based learning methods distilled from GPT. To ensure a fair comparison, we incorporate the same time series encoder (TSE) as used in our model for multimodal processing.

\item \textbf{ChatGPT}: ChatGPT \cite{open2023chatgpt} is a state-of-the-art LLM optimized for conversational applications, such as dialogue, summarization, and text completion.

\end{itemize}

\subsection{Implementation Details} \label{Implementation Details}
For our experiments, we utilized version 2.0.1 of the PyTorch framework, running on a CUDA 11.7 setup. The training processes were conducted using the DeepSpeed\footnote{https://github.com/microsoft/DeepSpeed} framework. We opted for the AdamW optimizer, starting with a learning rate of $1e^{-5}$ and incorporating a weight decay of 0.05. We implemented a warm-up phase that spanned 10\% of the training period. The experimental setup included two NVIDIA A100 GPUs, each with 80 GB of memory. To process time series data consistently, we padded all lab test results to a standard length of 1,000 time steps, dividing the data into 125 patches, where each patch included 8-time steps.

\subsection{Rationale Evaluation Metrics}

We defined the rationale evaluation metrics for the LLM and human evaluation as follows:  1). \emph{Correctness:} how medically accurate the rationale supports the diagnosis results. 2). \emph{Readability:} the extent to which a clinical rationale adheres to proper grammar and structural rules. 3). \emph{Soundness:} the logical coherence and insight provided by the clinical rationale. 4). \emph{Consistency:} the degree of alignment between the clinical rationale derived from medical notes and lab test results. 5). \emph{Persuasiveness:} the effectiveness of the clinical rationale in convincing the reader of its validity.

Evaluation scores based on Likert scale:
\begin{enumerate}
\item Strongly disagree
\item Disagree
\item Neither agree nor disagree
\item Agree
\item Strongly agree
\end{enumerate}

\subsection{Discussion on Different SLMs and LLMs}
In this section, we discuss the influence of different choices of base SLMs and teacher LLMs for our research, as presented in Table~\ref{table_appendix}. There are several findings: 1). \textbf{Decoder-Only SLMs Benefit More}: Decoder-only models, such as OPT\cite{zhang2022opt} and QWEN-2.5\cite{qwen2025qwen25technicalreport}, gain an average improvement of 3\% over Clinical-COT (single-modal CoT distillation) when using our multimodal CoT distillation method. This boost is attributed to their fully autoregressive nature, which integrates multimodal inputs and outputs within a single processing pipeline, effectively capturing implicit cross-modal dependencies. 2). \textbf{Encoder-Decoder SLMs Achieve Better Overall Performance}:
Encoder-decoder models handle longer input sequences more efficiently, since they do not allocate input space to labels, leading to improved disease diagnosis performance, particularly in environments with constrained computing resources. 3). \textbf{Comparable Performance Across Teacher LLMs}: SLMs distilled from DeepSeek and ChatGPT yield similar micro and macro F1 scores in disease diagnosis, underscoring the generality and stability of our approach. This consistency is achieved through carefully designed few-shot instruction templates for LLMs.

Overall, these results indicate that our multimodal CoT distillation mechanism ClinRaGen is a robust and generic method for enhancing diverse SLMs to achieve performance comparable to LLMs in healthcare tasks.

\end{document}